%% file: main.tex
\documentclass{article}

\usepackage[final,dblblindworkshop]{neurips_2026}
\workshoptitle{AI for Science: Verification in the Age of AI Scientists}

\makeatletter
\renewcommand{\@noticestring}{Accepted at the AI for Science workshop (NeurIPS 2026).}
\makeatother

\input{math_commands.tex}

\usepackage[utf8]{inputenc}
\usepackage[T1]{fontenc}
\usepackage{hyperref}
\hypersetup{pdfborder={0 0 1},citebordercolor={0 1 0},linkbordercolor={1 0 0},urlbordercolor={0 1 1}}
\usepackage{url}
\usepackage{booktabs}
\usepackage{array}
\usepackage{siunitx}
\usepackage{amsmath,amsfonts,amssymb}
\usepackage{graphicx}
\usepackage{multirow}
\usepackage{xcolor}
\usepackage{enumitem}


\title{Prior or Feedback? What an LLM Uses When Adapting Neural Operators}

\author{%
  Julian Chan\thanks{Equal contribution.} \\
  University of Surrey \\
  \And
  Javier Mora Jimenez\footnotemark[1] \\
  Compare the Market \\
}

\begin{document}
\maketitle

\begin{abstract}
Do LLM scientific agents rely only on their initial task context, or do they adapt their decisions in response to experimental feedback? We study this question in neural operator adaptation, where a large language model (LLM) selects fine-tuning configurations under a limited trial budget. Across transfers within and between partial differential equation (PDE) families, the LLM achieves lower held-out test nRMSE than random search and Bayesian optimisation in nearly every matched comparison.
Endpoint performance alone cannot distinguish what happens, so we verify each attribution with controlled interventions. Before observing any validation score, the LLM's first configuration already ranks near the top of the corresponding random-search pool, indicating a useful initial bias. 
A complementary cold-start intervention shows that the selected base learning rate shifts with the PDE description.
Once feedback becomes available, reassigning validation scores among evaluated configurations changes the next proposal in every case tested, whereas a value-preserving rewrite produces no comparable aggregate effect. 
These interventions establish that the LLM's decision-level actions respond to the given task and observed outcomes, showing that it combines a task-dependent prior with sensitivity to experimental feedback.
\end{abstract}

\section{Introduction}
\label{sec:intro}

Scientific agents operate sequentially: they propose actions, observe outcomes, and decide what to try next. Recent LLM-based systems extend this loop to hypothesis generation and refinement~\citep{aiScientist2024,aiCoScientist2025}. As such systems run more autonomous experiments, trusting their conclusions increasingly requires checking whether their decisions respond to the evidence they collect. Once the sequence ends, a benchmark score measures overall performance but does not reveal which inputs shaped each decision.

We study this distinction in a bounded experimental loop where a frozen large language model (LLM) selects fine-tuning configurations for a pretrained neural operator. The LLM \emph{adaptation policy} can propose and revise configurations, but cannot alter the data, action space, trial budget, or external evaluation protocol. At each search trial, the selected configuration is executed and scored.

The LLM receives a natural-language description of a target partial differential equation (PDE), the history of previously evaluated configurations and validation scores, and training diagnostics. 
Before the first evaluation, only the problem description is available. We use \emph{cold-start prior} to denote the preference induced by the LLM policy at this stage. The term refers to policy behaviour and does not imply pretrained knowledge of the underlying physics.
After the first evaluation, subsequent proposals can additionally depend on validation feedback. We therefore ask whether the cold-start prior provides a useful starting point and whether later experimental outcomes causally change the policy's actions. 

We study these questions using a pretrained Fourier neural operator (FNO), a neural operator architecture for learning solution maps between function spaces~\citep{li2021fno,kovachki2023neural}. We fine-tune this FNO after the target physical regime shifts away from its pretraining distribution.
Such shifts can degrade surrogate accuracy, and recent work addresses them through model scaling or broader pretraining across PDE families~\mbox{\citep{mpp2023,poseidon2024,dpot2024,walrus2025}}.
We instead hold the pretrained checkpoint and target-regime data fixed and study how effectively different policies configure adaptation under a small experimental budget. This setting reflects applications in which target-regime data are costly to generate and only a limited number of fine-tuning attempts are feasible~\citep{carey2025dataefficiency,lunagutierrez2026hlmbo}. Best practices for adapting these models remain unsettled~\citep{pift2026,song2026unsupervised}, and configuring the adaptation calls for machine learning expertise that domain scientists often do not have~\citep{liu2026autosurrogate}.
At each trial, a policy selects the optimiser, schedule, base learning rate, parameter-block multipliers, weight decay, rollout length, data loss, and two auxiliary physics-loss weights, then fine-tunes a fresh copy of the FNO. Model size and epochs are fixed across policies and excluded from the action space, so performance differences arise only from configuration choices. Each policy receives 20 trials to identify settings that transfer well.

We compare policies with progressively richer information. Random search uses neither problem context nor previous outcomes, whereas a Tree-structured Parzen Estimator (TPE), which handles mixed categorical and continuous spaces, models the observed configuration--score history~\citep{tpe2011,optuna2019}. The LLM receives this same history together with the PDE description and training diagnostics. All three policies operate under the same 20-trial budget, but differ in their proposal mechanisms, computational cost, and available information. Endpoint performance can therefore show whether the LLM is competitive under this budget, but not which of its additional inputs influence its decisions. This motivates the controlled input interventions below.

We test which displayed inputs shape those decisions by intervening on one input at a time: at cold start we vary the displayed problem description, and once a history has accumulated we reassign validation scores among the evaluated configurations. Neither intervention retrains the policy or alters the downstream optimisation task.

\paragraph{Contributions.}
\begin{itemize}[leftmargin=1.5em,itemsep=0pt]
\item \textbf{Competitive optimisation under a 20-trial budget.}
With the same 20-trial budget, the LLM policy achieves lower held-out test nRMSE
than random search and than TPE, spanning
within- and cross-family transfer.

\item \textbf{A decision-level verification protocol.}
We intervene on one input at a time and read the effect as an
\emph{action distance} between proposed configurations. 

\item \textbf{A task-dependent prior and feedback-sensitive proposals.}
Switching the displayed PDE family doubles
the base learning rate in the first proposal. Reassigning validation scores has an effect of about one categorical-coordinate change ($1/12\approx0.083$).

\end{itemize}

\paragraph{Related work.}
LLMs have been used for Bayesian optimisation, from proposing configurations directly from solution--score histories to interacting with established optimisers through tools~\citep{opro2024,llmhpo2023,sllmbo2024,llambo2024,bopro2025,brunzema2026agentic}. Some of these works read the effect of the problem description on search performance, whereas we also read its effect on the next proposed configuration. Recent matched comparisons show that LLM advantages are not universal: early gains can reflect default configurations, classical optimisers can outperform standalone LLM proposals, and agent gains vary across repository-level tasks~\citep{rodrigues2026llmworth,ferreira2026llmsbeat,huai2026agenthpobench}. No policy in our study starts from a default, so our comparison is specific to operator fine-tuning under a fixed budget.

Closest to our question, \citet{redko2026prior} study prior effects in hardware-aware code optimisation. We instead manipulate the displayed PDE description and validation history. This design relates to permuted-feedback controls, causal perturbations of stored experience, withheld feedback, and demonstration-label randomisation~\citep{wainrib2026aiagents,zhao2026faithful,huai2026agenthpobench,min2022demonstrations}. \citet{yagubyan2026consistent} measures agent reproducibility across repeated identical invocations using a distance over structured actions, and \citet{liao2026provenance} tests whether agents respond to the source of evidence under a matched intervention. We combine the two designs, so our intervention returns the size of the change rather than an indicator that it changed. PDE agents have also been used to construct PINN workflows, numerical solvers, and operator-inference models~\citep{pinnsagent2025,langpinn2025,codepde2025,opinflmm2026}; we instead hold the FNO architecture, downstream task, and action space fixed.

\begin{figure}[t]
\centering
\includegraphics[width=\linewidth]{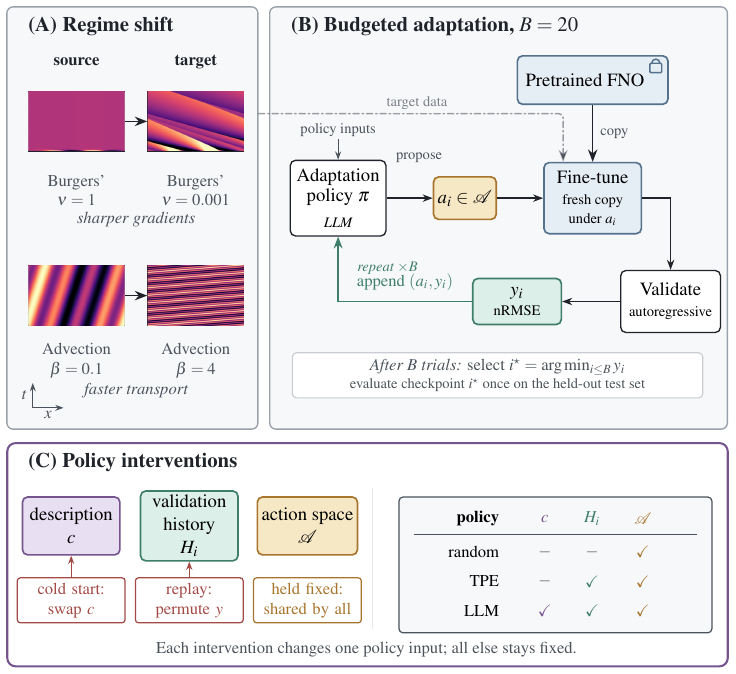}
\caption{\textbf{Budgeted FNO adaptation and controlled input interventions.}
\textbf{(A)} Within-family PDEBench shifts used as targets. Lower viscosity
$\nu$ sharpens Burgers' gradients, whereas higher advection speed $\beta$
increases the transport rate. \textbf{(B)} At each of $B=20$ trials, a policy
selects $a_i\in\mathcal A$, a fresh copy of the pretrained FNO is fine-tuned,
and its normalised root mean squared error (nRMSE) on the autoregressive
validation rollout is appended to $H_i$. After search, the configuration
selected by validation nRMSE is evaluated once on the held-out test set.
\textbf{(C)} At cold start, the displayed PDE description is changed; after a
history has accumulated, replay interventions reassign validation scores in
$H_i$. The FNO, data, compute budget, and $\mathcal A$ remain fixed.}
\label{fig:method}
\end{figure}

\section{Experimental setting: budgeted FNO adaptation}
\label{sec:setup}

We study budgeted adaptation using public PDEBench trajectories and pretrained
FNO checkpoints across two PDE families, each spanning multiple parameter
regimes~\citep{pdebench2022,pdebenchdata,pdebenchmodels}. Within each matched
comparison only the adaptation policy varies
(Appendices~\ref{app:fno}--\ref{app:shared-action-space}).

\paragraph{Adaptation problem.}
A run adapts one pretrained \emph{source} checkpoint to a \emph{target regime}
over a budget of $B=20$ sequential evaluations. At trial $i$, policy $\pi$
selects a configuration $a_i\in\mathcal A$ using the preceding
configuration--score history
\[
H_i=\{(a_j,y_j)\}_{j<i},
\]
together with any policy-specific inputs. Fine-tuning a fresh copy of the source
checkpoint under $a_i$ produces validation score $y_i$, which is appended to
the history before the next trial. The LLM policy also receives the
natural-language PDE description, denoted by $c$, and training diagnostics
(Figure~\ref{fig:method}B).

\paragraph{Target PDE families.}
We consider two families of one-dimensional scalar PDEs. The advection equation,
\[
\partial_t u+\beta\,\partial_xu=0,
\]
linearly transports the input profile at speed $\beta$. Burgers' equation adds
nonlinear self-advection and diffusion,
\[
\partial_t u+\partial_x(u^2/2)
=
\frac{\nu}{\pi}\,\partial_{xx}u.
\]
Reducing the viscosity $\nu$ produces sharper gradients and increases the
sensitivity of autoregressive rollouts to local prediction error
(Figure~\ref{fig:method}A).

\paragraph{Adaptation cells.}
An adaptation \emph{cell} pairs a source checkpoint with a target regime.
The six within-family cells change the governing parameter while preserving the
equation: advection $\beta=0.1$ is adapted to
$\beta\in\{0.4,1.0,4.0\}$, and Burgers' $\nu=1.0$ is adapted to
$\nu\in\{0.1,0.01,0.001\}$. Three cross-family cells adapt the advection
checkpoint to the Burgers' targets, and three further cells reverse that
transfer. We evaluate every cell with
three random-number seeds that resample the train--validation split, and call
each cell--seed pairing a \emph{trace} (Table~\ref{tab:cell-families}).

\begin{table}[!b]
\caption{Adaptation families. Each source--target cell is evaluated with three
seeds.}
\label{tab:cell-families}
\begin{center}
\small
\begin{tabular}{lll}
\toprule
\textbf{Family} & \textbf{Source $\to$ target} & \textbf{Change} \\
\midrule
\multirow{2}{*}{Within} & Advection $\beta{=}0.1\to\beta{\in}\{0.4,1.0,4.0\}$ & speed \\
 & Burgers' $\nu{=}1.0\to\nu{\in}\{0.1,0.01,0.001\}$ & viscosity \\
\addlinespace
\multirow{2}{*}{Cross} & Advection $\beta{=}0.1\to$ Burgers' targets & equation \\
 & Burgers' $\nu{=}1.0\to$ advection targets & equation \\
\bottomrule
\end{tabular}
\end{center}
\end{table}

\paragraph{Shared action space, training, and evaluation.}
A configuration specifies the optimiser, learning-rate schedule, base learning rate,
parameter-block learning-rate multipliers, weight decay, rollout length, data
loss, and two auxiliary physics-loss weights. Each trial begins from a fresh
copy of the source checkpoint, trains on $750$ trajectories, and computes
validation nRMSE against the run's fixed $350$-trajectory validation
split.

The first $1{,}000$ PDEBench trajectories form a held-out test prefix, on which the configuration with the lowest
validation nRMSE is evaluated once after the $20$-trial search. This prefix is
PDEBench's own test split, so the released checkpoints did not train on the
trajectories used to report held-out test nRMSE.

\paragraph{Policy implementations.}
Our main TPE configuration uses five random startup trials rather
than Optuna's default ten, leaving fifteen model-based proposals within the
$20$-trial budget. A separate control restores the default ten and extends
TPE's budget to $25$ evaluations, so that it keeps all fifteen model-based
proposals.

The LLM policy uses the hosted model
\texttt{deepseek-v4-pro}, with thinking and JSON modes enabled at
\texttt{reasoning\_effort=max}~\citep{deepseekThinkingMode,deepseekV4Preview};
Appendix~\ref{app:llm-config} pins the served model and the sampling settings.
At each trial, the LLM samples $S=3$ JSON configurations and executes the
medoid, the sampled configuration with the smallest total distance to the
others under Equation~\ref{eq:distance}. The rule limits the influence of a
single atypical completion but cannot correct a bias shared by all three, so
our behavioural claims concern that chosen configuration.

\section{Controlled input interventions}
\label{sec:tests}

Our decision-level verification protocol intervenes on one policy input at a
time while holding the FNO, data, trial budget, and action space fixed.

\paragraph{Action distance.}
Let $\mathcal C$ index the eight normalised numeric coordinates and
$\mathcal K$ the four categorical coordinates, with $D=12$. Both interventions
measure proposal displacement using
\begin{equation}
d(a,a') = \frac{1}{D}\left(
\sum_{j\in\mathcal C}\left\lvert\bar{a}_j-\bar{a}'_j\right\rvert
+
\sum_{j\in\mathcal K}\mathbb{1}\!\left[a_j\neq a'_j\right]
\right),
\label{eq:distance}
\end{equation}
where $\bar a_j$ denotes numeric coordinate $j$ rescaled to $[0,1]$.
Numeric coordinates contribute their absolute normalised difference, whereas
categorical coordinates contribute zero for a match and one for a mismatch.
Appendix~\ref{app:shared-action-space} gives the coordinate rescalings.
A change in one categorical coordinate contributes
$1/12\approx0.083$. We report distances in this unit, and set a decision
threshold to roughly $60\%$ of one such change (0.05).

\paragraph{Cold-start description test.}
In the advection $\beta=4.0$ and Burgers' $\nu=0.001$ cells, we show one of
three texts before the first trial: an advection description, a Burgers'
description, or none.
Appendix~\ref{app:coldstart-prompts} gives the exact prompts.
The analysis tests a directional
shift in the base learning rate of the first proposal.

\paragraph{Feedback replay and decision rule.}
For within-family cells, replay reconstructs a logged decision point and regenerates the next proposal
without continuing the search. The reassignment intervention preserves the
evaluated configurations and validation scores but permutes their
assignments. A notation control instead preserves every score value and its
assignment, rewriting each score in the equivalent scientific notation, for
example $0.0129$ as $1.29\mathrm{e}{-}2$.

For each trace, we
replay one decision point in each of three matched windows spanning the search,
measuring the action distance from the logged configuration to the regenerated
proposal. The trace effect is the median intervention-minus-control difference
in that distance across the three windows (Appendix~\ref{app:replay-effect}). We then take the median
trace effect across three seeds to obtain one effect for each of the six cells. Windows and resamples are repeated measurements within a trace, so the test treats each cell, not each decision, as an independent unit. An exact one-sided sign-flip test over these six cell effects has a minimum attainable $p$-value of $2^{-6}=1/64=0.015625$.

\begin{table}[!b]
\caption{Median held-out test nRMSE across three matched seeds after 20
evaluations; lower is better and bold marks the lowest median. $\dagger$ TPE beats
the LLM for seed 43 in this cell ($0.0263$ versus $0.0346$); the LLM wins the
other two seeds. In-domain is a released PDEBench FNO trained on the target
regime with about $9{,}000$ trajectories, against $750$ per adaptation trial;
it is a reference scale, not a matched baseline, and is not bolded.}
\label{tab:absolute-endpoint}
\centering
\small
\setlength{\tabcolsep}{9pt}
\begin{tabular}{@{}l *{4}{S[table-format=1.5,detect-weight=true,group-digits=false]}@{}}
\toprule
\textbf{Target cell} & {\textbf{LLM}} & {\textbf{Random}} & {\textbf{TPE}} & {\textbf{In-domain}} \\
\midrule
\multicolumn{5}{@{}l}{\emph{Within-family}} \\
Adv., $\beta=0.4$ & {\bfseries 0.01220} & 0.01750 & 0.01530 & 0.01330 \\
Adv., $\beta=1$ & {\bfseries 0.01340} & 0.01810 & 0.01660 & 0.01280 \\
Adv., $\beta=4$ & {\bfseries 0.00792} & 0.01580 & 0.00901 & 0.00880 \\
Burg., $\nu=0.001$ & {\bfseries 0.01970} & 0.06990 & 0.05220 & 0.03900 \\
Burg., $\nu=0.01$ & {\bfseries 0.00409} & 0.01580 & 0.00981 & 0.01040 \\
Burg., $\nu=0.1$ & {\bfseries 0.00155} & 0.00320 & 0.00248 & 0.00390 \\
\midrule
\multicolumn{5}{@{}l}{\emph{Cross-family}} \\
Adv.$\to$Burg., $\nu=0.001$ & {\bfseries 0.01990} & 0.07670 & {0.04010\rlap{$^\dagger$}} & 0.03900 \\
Adv.$\to$Burg., $\nu=0.01$ & {\bfseries 0.00350} & 0.01360 & 0.00709 & 0.01040 \\
Adv.$\to$Burg., $\nu=0.1$ & {\bfseries 0.00160} & 0.00287 & 0.00180 & 0.00390 \\
\midrule
\multicolumn{5}{@{}l}{\emph{Reverse cross-family}} \\
Burg.$\to$Adv., $\beta=0.4$ & {\bfseries 0.01280} & 0.02030 & 0.01520 & 0.01330 \\
Burg.$\to$Adv., $\beta=1$ & {\bfseries 0.01180} & 0.01930 & 0.01510 & 0.01280 \\
Burg.$\to$Adv., $\beta=4$ & {\bfseries 0.00707} & 0.01230 & 0.00934 & 0.00880 \\
\bottomrule
\end{tabular}
\end{table}

\section{Results}
\label{sec:results}

We first compare endpoint performance under the shared trial budget, then use controlled interventions to determine whether the PDE description and validation feedback shape the LLM's proposals.

\subsection{The LLM is competitive and starts from a useful proposal}
\label{sec:performance}

Under the shared 20-trial budget, the LLM policy finds a configuration
with lower held-out test nRMSE than random search in all $36$ matched runs and than TPE
in $35$ of $36$ (Table~\ref{tab:absolute-endpoint}). All $12$ cell medians
favour the LLM, across both within- and cross-family transfer.

This is a finite-budget advantage, not evidence of asymptotic superiority: TPE
is still improving at trial $20$ in several advection-to-Burgers' runs
(Figure~\ref{fig:supp-traces}). Giving TPE $25$ evaluations in a separate two-cell control still leaves the
$20$-trial LLM with lower held-out test nRMSE (Appendix~\ref{app:tpe-startup}), suggesting that the ranking 
does not arise solely from TPE's startup allocation.

Before observing any validation score, the median first LLM
proposal has lower validation nRMSE than $91.7\%$ of the $60$ random-search
configurations evaluated in the corresponding cell; only two of the $36$ first
proposals fall below their cell's random median. This indicates that the LLM enters the
search in a useful region before receiving feedback
(Appendix~\ref{app:performance}).

\begin{figure}[!b]
\centering
\includegraphics[width=\linewidth]{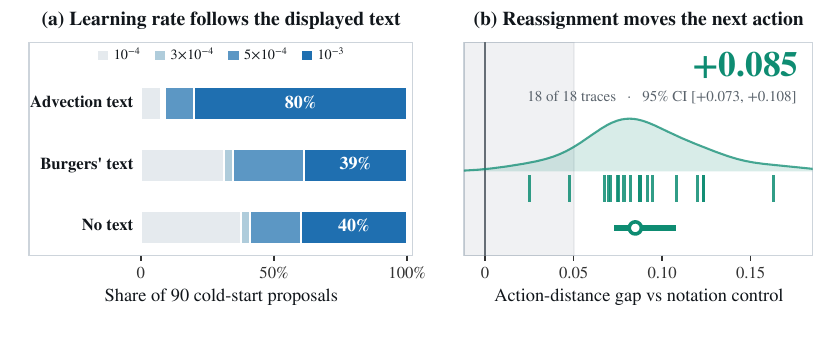}
\caption{Controlled input interventions. \textbf{(a)} Distribution of the base
learning rate over $90$ cold-start proposals per displayed description, pooling
$45$ from each target cell because the same text gives near-identical
distributions in both; labels show the share at $0.001$. \textbf{(b)} Reassignment-minus-notation action
distance across $18$ within-family traces. The marker gives the median and
bootstrap $95\%$ interval; shading marks values below the $0.05$ decision
threshold.}
\label{fig:interventions}
\end{figure}

\subsection{The PDE description shifts the base learning rate}
\label{sec:language}

Before any validation feedback, we show an advection description, a Burgers'
description, or no PDE description in both target cells. Each condition yields
$90$ cold-start proposals, with $45$ proposals per cell.

Switching the displayed description from Burgers' to advection doubles the
median base learning rate in the first proposal, from $0.0005$ to $0.001$, in
both target cells (Figure~\ref{fig:interventions}(a)). A one-sided
Mann--Whitney test over the pooled proposals (90 per displayed description, 45
from each target cell) gives $p=5.2\times10^{-9}$
(Appendix~\ref{app:coldstart}).
The no-description condition has a median base learning rate of $0.0005$ in
both cells. The advection description selects $0.001$ in $80\%$ of proposals,
compared with $39\%$ under the Burgers' description and $40\%$ without a
description. Changing only the displayed text therefore changes the initial
proposal, although the detectable effect is confined to one of twelve
coordinates.

\subsection{Reassigning validation scores changes the next configuration}
\label{sec:permutation}

We reassign the observed validation scores among previously evaluated
configurations and compare the resulting action with a notation control
that preserves every score and assignment.

Reassignment increases the median trace effect by $+0.085$ relative to this
control (bootstrap $95\%$ interval $[+0.073,+0.108]$;
Figure~\ref{fig:interventions}(b)), close to the $1/12=0.083$ contribution of
one categorical-coordinate change. All $18$ trace effects and six cell medians
are positive, exceeding our $0.05$ decision threshold with
$p=0.015625$. The notation control instead has median effect zero relative to
identical-prompt resampling. This effect persists when each score is reassigned together with the
diagnostics from its own trial (Appendix~\ref{app:feedback-bundle}). The next proposal therefore depends on which
outcomes are assigned to which configurations, not merely on formatting or
repeated sampling.

The response is also selective: reassignment changes rollout length in
$49.6\%$ of proposals, data loss in $40.4\%$, and schedule in $25.9\%$,
compared with $14.4\%$, $15.9\%$, and $14.1\%$ under resampling. Optimiser
changes remain near the resampling rate. We treat this coordinate ranking as
descriptive because we apply no per-coordinate tests. The policy therefore uses
the displayed feedback when selecting its next action, but the intervention
does not show that this response is rational or improves held-out test
performance. Appendix Tables~\ref{tab:permutation}
and~\ref{tab:replay-flips} give the complete contrasts.

\section{Discussion}
\label{sec:discussion}

The endpoint advantage under the fixed budget
(Table~\ref{tab:absolute-endpoint}) positions the LLM policy as a complement to
broader operator pretraining
\citep{mpp2023,poseidon2024,dpot2024,walrus2025} and LLM-generated scientific
solvers \citep{pinnsagent2025,langpinn2025,codepde2025,opinflmm2026}. Those
approaches build generality into the operator weights or generated code; ours
instead leaves both untouched and puts the expertise in the loop: the LLM
draws actions from a fixed action space and configures each fine-tuning trial. The architecture
and starting checkpoint never change, only how the operator adapts. This is useful when target-regime data,
training runs, or the expertise to run them are limited.

Beyond endpoint performance, our verification protocol tests whether the policy
uses its two distinctive inputs: the task description and observed feedback.
Random search conditions on neither, while TPE conditions only on numeric
history. Swapping the displayed PDE family shifts the median base learning rate of the LLM's first proposal, 
while permuting scores across evaluated configurations changes its next proposal by about one categorical coordinate (Figure~\ref{fig:interventions}), primarily through rollout length and data loss. 
Together, these interventions show that the LLM uses the task description to form its cold-start prior and observed feedback to guide later proposals.
The protocol requires only logged decision points and a distance over the action space,
so it applies to hosted LLMs without access to their internals, contributing to other decision-level checks of
AI agents
\citep{wainrib2026aiagents,zhao2026faithful,yagubyan2026consistent,liao2026provenance}.

Decision-level verification matters in unattended experimental loops, where two agents may reach the same endpoint score even if only one uses the evidence it gathers. 
The interventions therefore support a narrow conclusion: the LLM policy depends on the task description and observed scores. 
They do not explain why the policy responds as it does, establish that its decisions are rational or reflect physical understanding, or show that either dependence explains the endpoint advantage. Our study is limited in scope to one LLM and two one-dimensional PDE families. Applying the protocol across model families, scientific domains, higher-dimensional PDEs, and larger budgets would test whether these sensitivities generalise and help the agent select better solutions.

\section{Conclusion}
\label{sec:conclusion}

Prior or feedback? In this fixed-budget adaptation loop, the LLM policy uses both: 
it starts from a useful prior shaped by the task description and adjusts its later decisions in response to feedback. 
Endpoint evaluation measures what the experimental loop produces, while controlled interventions test what its decisions depend on. 
We argue that scientific agent verification requires both.

\clearpage
\renewcommand{\acksection}{\section*{Acknowledgments}}
\begin{ack}
We thank Niall Adams (Imperial College London) for comments on a draft of
this paper. Experiments were run on the Eureka2 computing cluster at the University of
Surrey.
\end{ack}

\bibliographystyle{plainnat}
\bibliography{refs}

\clearpage
\appendix

\section{Setup and reproducibility}
\label{app:reproducibility}

This appendix gives the checkpoints, training setup, action space, LLM
configuration, and compute behind the three main comparisons, together with the controls and protocol
deviations for each reported effect.
The search code is available at \url{https://github.com/julian-8897/budgeted-search-ai4science}. The run records
behind the reported results, with a script that recomputes them, are archived
at \href{https://doi.org/10.5281/zenodo.23213042}{doi:10.5281/zenodo.23213042}.

\subsection{Pretrained checkpoints and architecture}
\label{app:fno}

The checkpoints share one FNO backbone and differ only in pretraining data. The
operator uses ten input frames, 12 Fourier modes, width 20, and four spectral
layers; PDEBench fields are downsampled from 1024 to 256 spatial points and from
201 to 41 snapshots. The released checkpoints were trained for $500$ epochs on
about $9{,}000$ trajectories per regime~\citep{pdebench2022,pdebenchmodels}, whereas each
adaptation trial fine-tunes a fresh copy for at most $100$ epochs on $750$
trajectories.

\subsection{Training and evaluation}
\label{app:training}

The rollout length $k$ sets how many autoregressive steps the training loss
covers. At each step the FNO's own prediction is fed back as its next input.
Setting $k=1$ recovers single-step training, and larger $k$ penalises the
compounding error that the autoregressive evaluation measures. Rollout length
changes training alone: every configuration is scored on the same autoregressive
rollout to the end of the trajectory, so validation and test values are
comparable across trials.

Two auxiliary loss terms let a configuration impose physical structure during
fine-tuning, and the policy sets their weights. Neither term takes a spatial
derivative, so both stay well posed on the coarse grid. The conservation proxy
penalises the squared change in the discrete total mass $\sum_x u$ between
consecutive steps. Both target families conserve $\int u\,\mathrm{d}x$ under
periodic boundaries, so the PDEBench reference trajectories incur almost no
penalty on this term. The semigroup proxy feeds the FNO its own one-step
prediction back as input and penalises $\lVert f(f(u_t))-u_{t+2}\rVert^2$
against the field two steps ahead.

\subsection{Action space and rescalings}
\label{app:shared-action-space}

The FNO is divided into four parameter blocks: lift, spectral, bypass, and
projection. Each carries its own learning rate
\(\eta_{\mathrm{block}}=\eta_{\mathrm{base}}m_{\mathrm{block}}\), so a
configuration sets one base rate and four multipliers rather than four
independent rates. Random search and TPE sample the base learning rate and the
four multipliers log-uniformly over the bounds in
Table~\ref{tab:action-bounds}.

To compute action distance, the base learning rate and four block multipliers
use \(\bar x=(\log_{10}x-\log_{10}x_{\min})/
(\log_{10}x_{\max}-\log_{10}x_{\min})\). Weight decay uses
\(\bar w=\log(1+1000w)/\log(1+1000w_{\max})\), with
\(w_{\max}=10^{-2}\). We use this form rather than the logarithmic rescaling
above because the weight-decay bound includes zero. The two proxy weights use
linear min--max scaling. Values are clipped to their bounds.
Equation~\ref{eq:distance} then combines absolute numeric differences with
indicators of categorical mismatches.

\begin{table}[ht]
\caption{Action dimensions and bounds shared by all policies. The four block
multipliers are separate coordinates. Together, the rows define eight numeric
and four categorical coordinates, matching $D=12$ in
Equation~\ref{eq:distance}.}
\label{tab:action-bounds}
\begin{center}
\small
\begin{tabular}{ll}
\toprule
\textbf{Dimension} & \textbf{Range} \\
\midrule
optimiser & \{AdamW, Adam\} \\
schedule & \{none, cosine, step\} \\
base learning rate $\eta_{\mathrm{base}}$ & $[10^{-5}, 10^{-2}]$ \\
block multipliers $m_{\mathrm{lift}},m_{\mathrm{spec}},m_{\mathrm{byp}},m_{\mathrm{proj}}$ (4) & $[0.5, 2.0]$ \\
weight decay & $[0, 10^{-2}]$ \\
rollout length $k$ & $\{1,2,4,8\}$ \\
data loss & \{MSE, relative $L_2$, MSE $+$ relative $L_2$\} \\
conservation-proxy weight & $[0, 1]$ \\
semigroup-proxy weight & $[0, 1]$ \\
\bottomrule
\end{tabular}
\end{center}
\end{table}

\subsection{LLM policy configuration}
\label{app:llm-config}

\paragraph{Served model.}
Every reported API call used the provider identifier \texttt{deepseek-v4-pro}
in July and early August 2026. The logs record one served
model name and one system fingerprint on every proposal:
\texttt{fp\_9954b31ca7\_prod0820\_fp8\_kvcache\_20260402}, and our conclusions concern the policy served during this interval.

\paragraph{Sampling parameters.}
Calls set \texttt{reasoning\_effort=max},
with thinking and JSON modes enabled. Thinking mode ignores the sampling
parameters, so no temperature or nucleus setting shaped the
completions~\citep{deepseekThinkingMode}. Proposal diversity comes instead from
the model's own non-determinism at \texttt{reasoning\_effort=max}, which is what
the identical-prompt resampling control in Appendix~\ref{app:replay-results}
measures.

\paragraph{Prompt composition.}
The system prompt supplied the displayed PDE description $c$, the data sizes,
the FNO architecture, the search protocol, and the action bounds in
Table~\ref{tab:action-bounds}. From the second trial onwards, the prompt added the
preceding configuration, its validation score, and the trial's training
diagnostics (its loss and proxy curves), together with the best score so far,
the signed improvement, the remaining
budget, and a condensed history of earlier trials.

\subsection{Experiment counts and compute}
\label{app:experiment-counts}

The reported experiments comprise $3{,}420$ completed policy decisions and
$10{,}260$ valid LLM-generated configurations under the $S=3$ policy. We ran each fine-tuning trial on multiple \texttt{2g.20gb} Multi-Instance GPU (MIG)
slices, about one quarter of an NVIDIA A100 80GB PCIe GPU.

\section{Feedback replay analysis}
\label{app:permutation}

\subsection{Replay procedure and aggregation}
\label{app:replay-effect}

Replay measures how an edited validation history changes the policy's next
$S=3$ medoid action. Each replay reconstructs the prompt
before a logged decision, applies the intervention, recomputes the derived
history summaries, and regenerates only the next proposal. The search itself is
never continued, so no fine-tuning trial is rerun.

A single replayed decision is repeated five times per condition, which leaves
four nested levels to reduce: resample, window, seed, and cell. The
$18$ traces are the independent units, and the
sign-flip test clusters those into the six cells; treating every replayed
decision as its own observation would understate the uncertainty rather than
change the size of the effect. For condition $q$, cell $g$, seed $s$, window $w$, and
resample $r$, we take medians at every level and difference the two conditions
in the middle:
\[
d_q(g,s,w,r)
\xrightarrow{\operatorname{med}_r}
m_q(g,s,w)
\xrightarrow{\text{reassignment}-\text{notation}}
\delta(g,s,w)
\xrightarrow{\operatorname{med}_w}
\Delta(g,s)
\xrightarrow{\operatorname{med}_s}
C_g.
\]
Reading left to right: the five resamples of one decision give one number per
window and condition; the two conditions are then differenced at the matched
window; and the three windows, three seeds, and six cells are collapsed in
turn. The reported effect is \(M=\operatorname{med}_{g,s}\Delta(g,s)\), and the
exact sign-flip statistic uses the six cell medians $C_g$.

Each of the 18 traces contributes three matched decision windows. For
every trace, a seeded draw selected one next-decision index
from each window: early in $\{2,3,4\}$, middle in $\{9,10,11\}$, and late in
$\{16,17,18\}$. Every reassignment is a non-identity permutation.
\subsection{Feedback replay results and controls}
\label{app:replay-results}

Both controls are defined against prompt resampling, to account for
the sampling variance of LLM generations. The notation control preserves every
score and its assignment while changing the characters, and the
cross-family control resamples from an identical prompt, so a gap measured
against either exceeds what repeated sampling alone produces.

Table~\ref{tab:permutation} separates the comparison of
reassignment against the notation control, the notation control itself, and the
test on cross-family transfers. The notation control is centred at zero
relative to identical-prompt resampling, with bootstrap 95\% interval
$[-0.006,+0.005]$, whereas reassignment relative to notation has interval
$[+0.073,+0.108]$. The cross-family replay shows a similar change in the next
action, but uses resampling rather than notation as its control.

The notation control edits the prompt more heavily than the reassignment does.
Rewriting each score in scientific notation moves a character-level Levenshtein
distance of $76$--$160$, against $22$--$95$ for reassigning the scores, and
still leaves the next proposal where identical-prompt resampling leaves it.

\begin{table}[!ht]
\caption{Action-distance summaries for the validation-history intervention.
Exact one-sided $p$-values use six cell medians. The cross-family result tests
whether the effect extends to cross-family transfers.}
\label{tab:permutation}
\begin{center}
\small
\setlength{\tabcolsep}{4pt}
\begin{tabular}{llrcc}
\toprule
\textbf{Context} & \textbf{Contrast} & \textbf{Traces} & \textbf{Median gap} & \textbf{$p$} \\
\midrule
Within-family & reassignment versus notation & 18 & $+0.085$ & $0.015625$ \\
Within-family & notation versus resampling & 18 & $+0.000$ & $0.63$ \\
Cross-family & reassignment versus resampling & 18 & $+0.083$ & $0.015625$ \\
\bottomrule
\end{tabular}
\end{center}
\end{table}

The breakdown by categorical coordinate in Table~\ref{tab:replay-flips} shows
that the response is concentrated in rollout length, data loss, and schedule.
The optimiser changes at close to its resampling rate, so reassignment leaves
some categorical choices alone. We apply no per-coordinate test, so we read this
ranking as a description of where the response falls and not as a test of it.

\begin{table}[!ht]
\caption{Within-family categorical flip rates relative to the logged configuration.
No per-coordinate tests are applied.}
\label{tab:replay-flips}
\begin{center}
\footnotesize
\begin{tabular}{lccc}
\toprule
\textbf{Coordinate} & \textbf{Resampling} & \textbf{Notation} & \textbf{Reassignment} \\
\midrule
Rollout length $k$ & $14.4\%$ & $11.1\%$ & $49.6\%$ \\
Data loss          & $15.9\%$ & $16.7\%$ & $40.4\%$ \\
Schedule           & $14.1\%$ & $13.3\%$ & $25.9\%$ \\
Optimiser          & $6.3\%$  & $7.0\%$  & $7.8\%$ \\
\bottomrule
\end{tabular}
\end{center}
\end{table}

\subsection{Feedback-bundle control}
\label{app:feedback-bundle}

Score reassignment leaves each trial's training diagnostics in place while
moving its score. In the
feedback-bundle condition, we permute each score together with its diagnostics
to preserve their pairing. The median action-distance gap relative to resampling is
$+0.080$ within-family and $+0.054$ cross-family, with $18/18$ positive traces
and exact one-sided sign-flip $p=0.015625$ in each.

\section{Cold-start description intervention}
\label{app:coldstart}

With no search history, we generate $n=45$ medoid proposals for each cell and
description condition. We show each cell the same advection and Burgers'
descriptions in separate prompts and omit the description in a third condition.

The statistic is the median $\log_{10}$ base learning rate per displayed
description, pooling the 45 proposals from each target cell. The 90
advection-description proposals are compared with the 90 Burgers'-description
proposals by a one-sided Mann--Whitney test in the pre-specified direction
($U=5777.5$ of $8100$ pairs). With only four distinct learning rates, the
null distribution can be enumerated exactly, giving $p=5.2\times10^{-9}$; the
tie-corrected normal approximation gives $p=1.0\times10^{-8}$. The median doubles in each cell individually, from
$0.0005$ to $0.001$.

Averaged over all twelve coordinates, the aggregate action distance between
descriptions is $+0.0025$, about one thirtieth of a single
categorical-coordinate change, with $p=0.015$ from a one-sided permutation test
that shuffles the description labels. An effect confined to one coordinate
shrinks by roughly this much when spread across the whole action vector. The
aggregate uses a different reference from replay, so its magnitude does not
compare with the replay effect in Section~\ref{sec:permutation}.

\subsection{Exact cold-start prompts}
\label{app:coldstart-prompts}

The prompt was written in Markdown and is reproduced below in LaTeX. Each description
appeared verbatim in both cells.

\begin{quote}
\footnotesize
\textbf{Adaptation}: 1D Advection $\beta{=}0.1 \to \beta{=}4.0$.
\[
\partial_t u + \beta\,\partial_x u = 0,
\quad x \in (0,1),
\quad t \in (0,2]
\]
\textbf{Source}: $\beta=0.1$ \;$\cdot$\;
\textbf{Target}: $\beta=4.0$ \;$\cdot$\;
\textbf{Pretrained checkpoint}: FNO trained at $\beta=0.1$ \;$\cdot$\;
\textbf{Exact solution}: $u(t,x)=u_0(x-\beta t)$
\end{quote}

\begin{quote}
\footnotesize
\textbf{Adaptation}: 1D Burgers' $\nu{=}1.0 \to \nu{=}0.001$.
\[
\partial_t u + \partial_x(u^2/2)
= \frac{\nu}{\pi}\,\partial_{xx} u,
\quad x \in (0,1),
\quad t \in (0,2]
\]
\textbf{Source}: $\nu=1.0$ \;$\cdot$\;
\textbf{Target}: $\nu=0.001$ \;$\cdot$\;
\textbf{Pretrained checkpoint}: FNO trained at $\nu=1.0$
\end{quote}

\section{Endpoint performance}
\label{app:performance}

\begin{figure}[!ht]
\centering
\includegraphics[width=\linewidth]{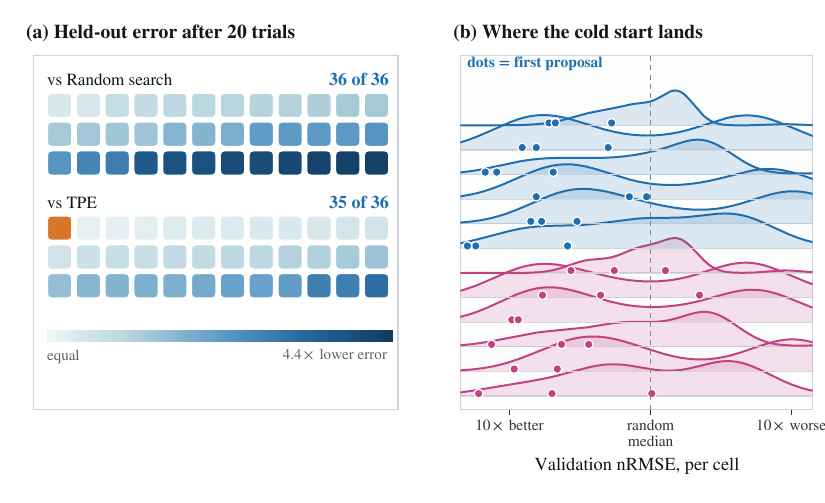}
\caption{Run-level endpoint performance and initial proposal quality.
\textbf{(a)} Paired baseline-to-LLM held-out test nRMSE ratios for each cell and
seed; values above one favour the LLM, and amber marks the single TPE win.
\textbf{(b)} Placement of each first proposal within the pooled distribution
of the $60$ random-search validation configurations from the same cell. Blue
denotes within-family and magenta cross-family transfer. Panel (b) does not
measure held-out test benefit.}
\label{fig:performance-start}
\end{figure}

We evaluate the configuration selected after the shared trial budget on the
held-out test set using nRMSE. Figure~\ref{fig:performance-start} shows the
run-level margins and the placement of each first proposal.

\paragraph{Placing the first proposal.}
Panel (b) of Figure~\ref{fig:performance-start} places each first proposal in
the pooled random-search distribution for its own cell. We measure that
placement as the share of the random pool scoring below the first proposal,
counting ties as half. For first-proposal score $x$ and the 60 pooled
random-search scores $R$ in the same cell, this lower-tail midrank is
\((\#\{R<x\}+\tfrac12\#\{R=x\})/60\). Across the 36 first proposals the
median lower-tail quantile is $8.3\%$, equivalent to beating $91.7\%$ of the
pool; 28 proposals beat at least $75\%$ of their pool, and 34 beat at least
half.

\paragraph{Zero-shot reference.}
Applying the source checkpoint to the target with no adaptation gives held-out
test nRMSE between $0.12$ and $1.24$ across the twelve cells, which is about
$5$--$400\times$ the worst of the three adapted policy medians in the same
cell. All three policies recover most of the error the regime shift causes, and
the comparisons above concern how well they do so within 20 trials.

\subsection{TPE with ten startup trials}
\label{app:tpe-startup}

The main comparison uses five startup trials, below Optuna's default of ten.
This control restores the default in one within-family and one cross-family
cell, with three seeds each. Ten startup trials would leave only ten
model-based proposals inside a 20-trial budget, so we extend TPE to 25 trials
to keep all fifteen, which gives TPE 25\% more evaluations than the LLM.

The LLM records lower held-out test nRMSE in 5/6 comparisons and both cell medians.
The single TPE win falls on the same cell and seed as in the main comparison,
where TPE reaches $0.0248$ here against $0.0263$ with five startup trials.
Median TPE-to-LLM ratios are $1.91\times$ within-family and $2.50\times$
cross-family.

\subsection{Search trajectories}
\label{app:traces}

\begin{figure}[p]
\centering
\textbf{Within-family cells (a)--(f)}\par\smallskip
\includegraphics[width=\linewidth]{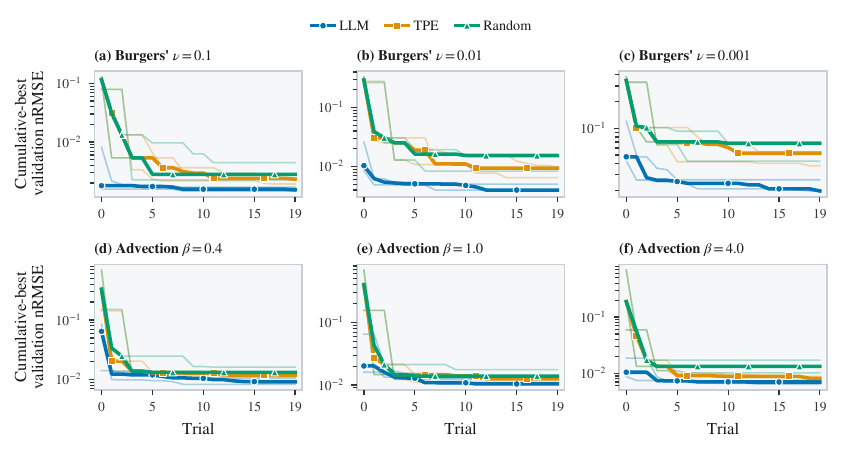}
\par\vspace{2mm}
\textbf{Cross-family cells (g)--(l)}\par\smallskip
\includegraphics[width=\linewidth]{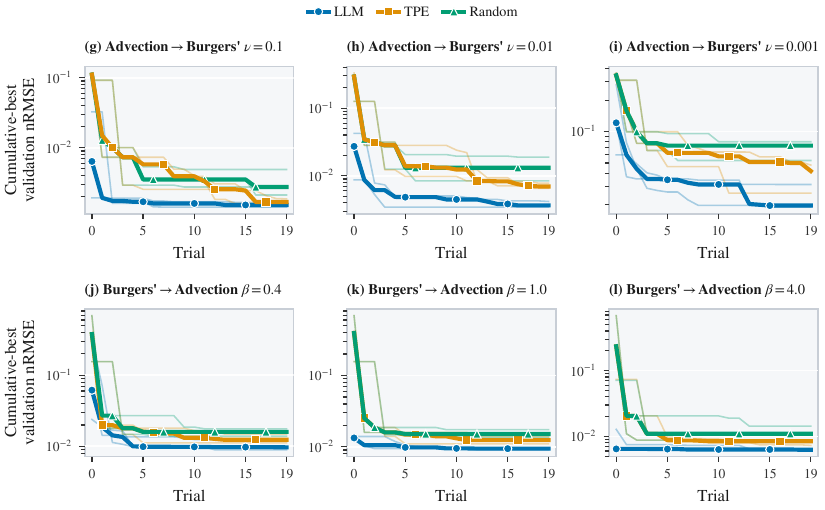}
\caption{Cumulative-best validation nRMSE as a function of trial number
for all cells, with three seeds per method. Thin lines are individual runs;
thick lines are medians across the three seeds. Panels
\textbf{(a)}--\textbf{(f)} are the within-family cells. Of the cross-family
panels, \textbf{(g)}--\textbf{(i)} adapt an advection-pretrained FNO to
Burgers' and \textbf{(j)}--\textbf{(l)} reverse the transfer.}
\label{fig:supp-traces}
\end{figure}

Figure~\ref{fig:supp-traces} shows how far validation error has fallen by each
trial, for every cell. Most improvements occur early for every method, and the
curves do not establish that any baseline has finished improving by trial
$20$.

\clearpage

\end{document}

%% file: math_commands.tex
\usepackage{amsmath,amsfonts,bm}

\def\eqref#1{equation~\ref{#1}}

\def\1{\bm{1}}

\DeclareMathAlphabet{\mathsfit}{\encodingdefault}{\sfdefault}{m}{sl}
\SetMathAlphabet{\mathsfit}{bold}{\encodingdefault}{\sfdefault}{bx}{n}